\documentclass[conference]{IEEEtran}
\usepackage{cite}
\usepackage{amsmath,amssymb,amsfonts}
\usepackage{algorithmic}
\usepackage{graphicx}
\usepackage{textcomp}
\usepackage{xcolor}
\usepackage{orcidlink}
\usepackage{hyperref}

\hypersetup{
    colorlinks=true,
    linkcolor=blue,
    filecolor=magenta,      
    urlcolor=blue,
}

\begin{document}

\title{HBRB-BoW: A Retrained Bag-of-Words Vocabulary for ORB-SLAM via Hierarchical BRB-KMeans}

\author{
    \IEEEauthorblockN{
        Minjae Lee~\orcidlink{0009-0005-2796-3207},
        Sang-Min Choi~\orcidlink{0000-0001-5950-3081},
        Gun-Woo Kim~\orcidlink{0000-0001-5643-4797},
        and Suwon Lee~\orcidlink{0000-0003-2603-1385}
    }
    \IEEEauthorblockA{
        Department of Computer Science and Engineering, Gyeongsang National University \\
        Republic of Korea \\
        Email: \{wjdchs0129, jerassi, gunwoo.kim, leesuwon\}@gnu.ac.kr
    }
}

\maketitle

\begin{abstract}
    In visual simultaneous localization and mapping (SLAM), the quality of the visual vocabulary is fundamental to the system's ability to represent environments and recognize locations. While ORB-SLAM is a widely used framework, its binary vocabulary, trained through the k-majority-based bag-of-words (BoW) approach, suffers from inherent precision loss. The inability of conventional binary clustering to represent subtle feature distributions leads to the degradation of visual words, a problem that is compounded as errors accumulate and propagate through the hierarchical tree structure. To address these structural deficiencies, this paper proposes hierarchical binary-to-real-and-back (HBRB)-BoW, a refined hierarchical binary vocabulary training algorithm. By integrating a global real-valued flow within the hierarchical clustering process, our method preserves high-fidelity descriptor information until the final binarization at the leaf nodes. Experimental results demonstrate that the proposed approach yields a more discriminative and well-structured vocabulary than traditional methods, significantly enhancing the representational integrity of the visual dictionary in complex environments. Furthermore, replacing the default ORB-SLAM vocabulary file with our HBRB-BoW file is expected to improve performance in loop closing and relocalization tasks.
\end{abstract}


\smallskip
\noindent \small \textbf{Code Availability:} The source code with retrained vocabulary file is available at \url{https://github.com/MinChoi0129/HBRB-BoW}.
\section{Introduction}
Simultaneous localization and mapping (SLAM) is an essential technology for realizing camera-based autonomous driving. Among non-deep learning-based SLAM methods, ORB-SLAM\cite{orbslam2, orbslam3} remains a significant tool in the field due to its proven performance. To address the problem of accumulated drift during the mapping process, most studies employ loop detection and loop closing techniques.

In visual SLAM systems, loop detection is primarily based on the bag-of-words (BoW) framework, where accurately recognizing previously visited locations by comparing the current view with a vast amount of past frame data is a critical factor for system stability. Specifically, ORB-SLAM utilizes the DBoW2 framework\cite{dbow} for visual vocabulary training. DBoW2 introduced support for binary descriptors and adopts a hierarchical tree structure to enhance search efficiency, relying on approximate nearest neighbor searches as its underlying mechanism.

The DBoW lineage has evolved through several iterations, including the original DBoW\footnote{\url{https://github.com/dorian3d/DBow}}, DBoW2\footnote{\url{https://github.com/dorian3d/DBoW2}}, DBoW3\footnote{\url{https://github.com/rmsalinas/DBow3}}, and FBoW\footnote{\url{https://github.com/rmsalinas/fbow}}. These advancements have predominantly focused on optimizing computational speed and memory efficiency. Despite these gains in throughput, they continue to rely on traditional binary clustering methods that lack the capacity to represent subtle feature distributions. This limitation leads to a cumulative loss of precision as quantization errors propagate through the hierarchical tree structure during the training process.

The vocabulary currently used in ORB-SLAM is trained on the Bovisa dataset \cite{bovisa}, using ORB \cite{orb} to detect keypoints and generate binary descriptors. While binary descriptors within the DBoW framework offer the distinct advantage of rapid revisit detection, they also entail several structural limitations.

\begin{table}[t]
    \centering
    \caption{Comparison of trajectory-based performance on the KITTI dataset between DBoW2 and HBRB-BoW.}
    \label{tab:performance_comparison}
    \renewcommand{\arraystretch}{1.2}
    \begin{tabular}{l|ccc}
        \hline
                           & DBoW2           & HBRB-BoW                 & $\Delta$        \\ \hline
        \multicolumn{4}{l}{\textbf{Translation [m]}}                                      \\ \hline
        ATE                & 8.140m          & \textbf{5.631m}          & 250.9cm         \\
        mRPE               & 5.063m          & \textbf{4.539m}          & 52.4cm          \\
        ATE (w/o seq. 19)  & \textbf{4.372m} & 4.435m                   & -6.3cm          \\
        mRPE (w/o seq. 19) & 4.414m          & \textbf{4.176m}          & 23.8cm          \\ \hline
        \multicolumn{4}{l}{\textbf{Rotation [$^\circ$]}}                                  \\\hline
        ATE                & 1.458$^{\circ}$ & \textbf{1.251}$^{\circ}$ & 0.207$^{\circ}$ \\
        mRPE               & 1.449$^{\circ}$ & \textbf{1.361}$^{\circ}$ & 0.088$^{\circ}$ \\
        ATE (w/o seq. 19)  & 1.224$^{\circ}$ & \textbf{1.137}$^{\circ}$ & 0.087$^{\circ}$ \\
        mRPE (w/o seq. 19) & 1.395$^{\circ}$ & \textbf{1.356}$^{\circ}$ & 0.039$^{\circ}$ \\ \hline
    \end{tabular}
\end{table}

\begin{figure*}[t]
    \centering
    \includegraphics[width=\linewidth]{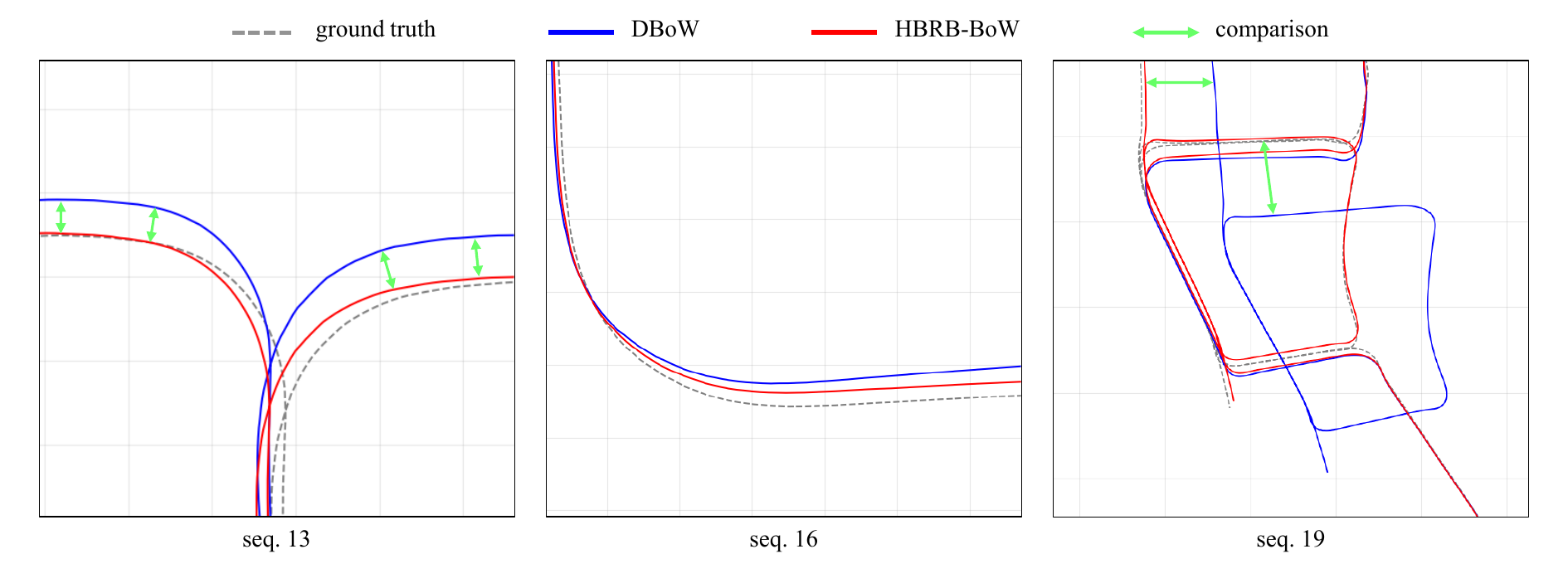}
    \caption{Qualitative comparison of trajectory examples.}
    \label{fig:qualitative}
\end{figure*}

First, during the clustering process for binary data, the k-majority method is adopted instead of the standard k-means to select centroids, and Hamming distance is used instead of Euclidean distance for measurement. Due to the inherent nature of binary space, which cannot represent decimal values, this approach inevitably leads to the loss of fine-grained information. Binary-to-real-and-back (BRB)-KMeans\cite{brbkmeans} addressed this issue by proposing a method that converts binary data into real-valued data to perform k-means and then restores it back to binary form, thereby contributing to improved clustering quality.

Second, DBoW performs clustering hierarchically. Due to the aforementioned information loss in binary clustering, errors originating at upper nodes accumulate and propagate to lower nodes. This propagation of errors eventually has a negative impact on the overall vocabulary training process.

Therefore, this study proposes hierarchical BRB-BoW (HBRB-BoW), a retraining algorithm designed to more effectively perform hierarchical binary vocabulary learning. Through various experiments, we demonstrate the effectiveness and utility of the proposed method.
\section{Hierarchical Binary Vocabulary Retraining}
Since BRB-KMeans assumes a non-hierarchical clustering scenario, it does not discuss the processing of tree structures. While leaf nodes, which serve as visual words, must ultimately take a binary form, two distinct approaches can be considered for the traversal from the root node to the leaf nodes.

The first approach is to perform clustering locally using the BRB method at every branching point. The second approach is to convert the data to a real-valued representation (binary-to-real) at the root node, maintain standard k-means in the real-valued domain throughout the hierarchy until reaching the leaf nodes, and finally restore it to binary form (real-to-binary) only at the leaf level. Through experimental validation, this study confirmed that the latter approach is superior in terms of performance and selected it as the proposed methodology.
\section{Experiments}

\subsection{Experimental Setup}
\label{sec:expsetup}
To ensure a fair comparison, the parameters for the vocabulary tree were set to a branching factor of $k = 10$ and a depth of $L = 6$, consistent with \cite{dbow}. Regarding the training dataset, although the Bovisa dataset consists of 65,047 images, \cite{dbow} utilized a subset of 10,000 images without specifying their detailed configuration.

To establish a reliable baseline, we performed random sampling of 10,000 images while varying the random seeds to train our own DBoW vocabularies. We identified the specific seed that resulted in a directly trained DBoW vocabulary with performance most consistent with the official vocabulary provided by the ORB-SLAM repository\footnote{\url{https://github.com/raulmur/ORB_SLAM2/tree/master/Vocabulary}}. This specific data subset was then employed to train our HBRB-BoW vocabulary, ensuring that the performance gains are attributable to the proposed algorithm rather than differences in training data.

The KITTI dataset was employed to evaluate the impact of the loop detection algorithm on overall system performance. Since the dataset provides only pose data as ground truth, we conducted a trajectory-based performance evaluation to indirectly assess loop detection effectiveness. For evaluation metrics, we utilized the absolute trajectory error (ATE) to measure global consistency and the relative pose error (RPE) to assess drift over specific distances. Specifically, the root mean squared error (RMSE) for RPE was measured at intervals from 100m to 800m, and their average, mRPE, is presented as the primary performance indicator. Note that sequences 01, 03, 04, 08, 10, 11, 12, 14, 17, 20, and 21, which lack loop closure occurrences, were excluded from the evaluation.

\subsection{Performance Comparison}
\label{sec:perf}
Table.~\ref{tab:performance_comparison} presents a comparison of trajectory-based performance using the original vocabulary (DBoW) bundled with ORB-SLAM and the HBRB-BoW vocabulary retrained using the proposed algorithm. As described in Section~\ref{sec:expsetup}, we substituted only the vocabulary within the same SLAM framework to isolate and analyze the performance impact of the visual dictionary itself.

The experimental results show that the system's performance improved significantly across all metrics when the HBRB-BoW vocabulary was employed. Notably, the translation ATE decreased from 8.140m to 5.631m, marking an improvement of approximately 30.8\%p and reducing the global trajectory error by 250.9cm. The mRPE also improved by 10.3\%p (from 5.063m to 4.539m), effectively suppressing cumulative drift by 52.4cm.

Furthermore, the superiority of the HBRB-BoW vocabulary remained evident even when excluding sequence 19, which exhibited anomalously high error across all models. In the analysis excluding sequence 19, the mRPE improved by 5.4\%p (from 4.414m to 4.176m), corresponding to an error reduction of 23.8cm. Rotation errors also showed a consistent decline in both ATE and mRPE. This indicates that the performance gains are not dependent on specific outliers but rather stem from the vocabulary's general ability to effectively stabilize trajectories across various driving environments.

In conclusion, the HBRB-BoW vocabulary provides more precise visual words by minimizing information loss throughout the hierarchical tree structure. This enhances the accuracy of loop closure detection, thereby ensuring the overall robustness of the SLAM system.

\subsection{Qualitative Study}
Fig.~\ref{fig:qualitative} provides a qualitative comparison demonstrating how effectively the HBRB-BoW vocabulary detects loops and produces superior trajectories across sequences in a real-world SLAM process. A closer alignment with the grey Ground Truth trajectory indicates higher estimation accuracy. It is evident that the vocabulary trained via the HBRB-BoW method produces trajectories significantly closer to the ground truth compared to the blue DBoW approach.

As discussed in Section~\ref{sec:perf}, the sequence 19 highlights the most substantial performance disparity. Visual inspection reveals that the DBoW fails to detect loops in this sequence, leaving cumulative drift uncorrected. In contrast, the HBRB-BoW vocabulary successfully identifies loop closure candidates, effectively eliminating accumulated errors and ensuring more stable trajectory estimation.
\section{Conclusion}
This paper proposed HBRB-BoW to address the precision loss in hierarchical binary vocabularies by preserving real-valued descriptor information until the final binarization at leaf nodes. Quantitative evaluations on the KITTI dataset demonstrate that the HBRB-BoW vocabulary significantly enhances system performance, reducing the translation ATE by 30.8\%p and the mRPE by 10.3\%p. Notably, our approach successfully resolved accumulated drift in challenging scenarios, such as sequence 19, where the baseline method failed to detect loop closures.

Ultimately, HBRB-BoW provides a more discriminative visual dictionary, ensuring the representational integrity and robustness of visual SLAM in complex environments. As our method is fully compatible with the existing framework, substantial improvements in loop closing and relocalization can be expected by simply substituting the original vocabulary file with the HBRB-BoW file.

\bibliographystyle{IEEEtran}
\bibliography{references}

\end{document}